# Mixing Metaphors

Mark G. Lee and John A. Barnden
School of Computer Science,
University of Birmingham
Birmingham, B15 2TT
United Kingdom
*mgl@cs.bham.ac.uk*
*jab@cs.bham.ac.uk*

**Abstract**

Mixed metaphors have been neglected in recent metaphor research. This paper suggests that such neglect is short-sighted. Though mixing is a more complex phenomenon than straight metaphors, the same kinds of reasoning and knowledge structures are required. This paper provides an analysis of both parallel and serial mixed metaphors within the framework of an AI system which is already capable of reasoning about straight metaphorical manifestations and argues that the processes underlying mixing are central to metaphorical meaning. Therefore, any theory of metaphors must be able to account for mixing.

## Introduction

The phenomenon of mixed metaphors has been largely neglected by previous research in metaphor understanding. This has been due to two prevalent assumptions. First, mixed metaphors are often regarded as examples of (at worst) pathological language use or (at best) poor style. Secondly, it is clear that the understanding of a mixed metaphor is more difficult that of a single metaphor, since a mix requires reasoning about several vehicle domains.

In this paper, we wish to argue that the former assumption is wrong: mixed metaphors are common in mundane everyday discourse and can be understood by hearers without recourse to specialised reasoning. In addition, the second assumption is detrimental to long term progress since mixed metaphorical manifestations rely on straight metaphors. More specifically, this paper makes the following claim: *the reasoning processes and data structures involved in understanding mixed metaphors are identical to those used in understanding straight metaphors*. Therefore, current research on metaphor processing should be capable of being extended to deal with mixed phenomena and mixing can provide valuable insight into the processes underlying straight metaphors.

To this end, this paper describes some initial work done with ATT-Meta [Barnden, 1997] to handle various types of mixing and reprises an earlier claim for the need for *within-vehicle* reasoning and the use of conversion rules to filter the relevant connotations of a particular metaphor.

Mixed metaphors are often regarded as humourous or cases of defective speech. Consider the following pathological sentence, quoted by Fowler [Fowler, 1908]:

1. *"This, as you know, was a burning question; and its unseasonable introduction threw a chill on the spirits of all our party."*

In example 1, the question is metaphorically "hot". However, its introduction makes the party's spirits "cold". Despite this contradiction, the sentence can be understood to mean that the question was somehow controversial and its inappropriate introduction saddened the emotions of the party members. Fowler criticised such examples as poor style.

However, despite the conflict between "hot" questions and "cold" emotions, the connotation of the sentence can be easily understood since it alludes to two well-known metaphors, i.e. "DIFFICULT QUESTIONS ARE HOT OBJECTS" and "SAD EMOTIONS ARE COLD OBJECTS". Furthermore, it is unlikely that most native speakers would even consider the disparity of "hot" questions causing "cold" reactions. This is because, in each case what is mapped is not an instance of temperature change, but *a connotation with direct relevance to the tenor domain*.

In this paper, we will argue that it is often necessary to do extended reasoning prior to mapping from vehicle to tenor. Therefore, a capacity for within-vehicle-reasoning is essential and any conversion must also act as a strict filter to limit the range of metaphorical meaning.

The paper is structured as follows: in Section 2, we will outline and distinguish two key types of mixed metaphor: serial and parallel. In Section 3, we will briefly outline ATT-Meta and provide an analysis of each type of mixed metaphor which our program is capable of dealing with and then in Section 4 extend the discussion to other types of mixes and the wider issues facing mixed metaphor research.

## 2. Mixed metaphor distinctions

It is possible to distinguish two types of mixed metaphor: parallel mixes and serial mixes. In a parallel mixed metaphor, the tenor (A) is seen as a vehicle (B) and then as a second vehicle (B'). In a serial mixed metaphor, the tenor (A) is seen as a vehicle (B) which is then seen as a different vehicle (C). The key distinction is that in parallel mixes the two metaphors do not metaphorically interact and in serial mixes, they do. For example, consider the following two mixed metaphors:

   2. *"The critique shed light on the theory's shakey foundations."*
   3. *"One part of John hotly contested the verdict."*

The utterance in 2 manifests two metaphors: "MENTAL INTERACTION AS VISION" and "THEORIES AS BUILDINGS" [Lakoff & Johnson, 1980; Grady, 1997]. However, what is novel is both metaphors are being applied in the same sentence. Following the definition given above for parallel mixes, the following domains are involved:

   A: Domain of theories, ideas, arguments etc.
   B: Domain of light/perception.
   B': Domain of buildings.

If ideas are light sources then a critique can shed light on a theory so that observers can "see" aspects of a theory. In this instance, the "light source" of the critique allows them to see the weak foundations of the theory. Weak foundations in a building suggest that the building itself might collapse, therefore, if theories are buildings then their weak foundations may cause the entire theory to collapse, or literally, be refuted. The above sentence can be unravelled by treating the different metaphorical vehicles separately since each applies to a different aspect of the tenor domain, i.e. the critique and the theory itself.

It is worth noting that the metaphor "MENTAL INTERACTION AS VISION" might itself be a combination of two familiar metaphors: "IDEAS ARE LIGHT SOURCES" [Lakoff & Johnson, 1980] and "UNDERSTANDING AS SEEING". We will return to this point in Section 4.

The utterance in example 3 also manifests two familiar metaphors: "MIND PARTS AS PERSONS" [Barnden,1997] (see also, Lakoff's metaphor "IDEAS ARE ENTITIES" [Lakoff, 1993]) and "ANGER IS HEAT" [Lakoff & Turner,1980]. In the "MIND PARTS AS PERSONS" metaphor, the mind is composed of different parts which may have different beliefs, intentions and personalities. Mentioning that "one part" of John contested the verdict suggests that there exists more than one part and that some other part of John did not contest the verdict. Moreover, the part of John referred to contested the verdict "hotly". In the "ANGER AS HEAT" metaphor, anger is seen as heat. Therefore, the "part" of John which contested the verdict did so with anger. Following the definitions given above, the following domains are involved:

   A: John's mental/emotional states.
   B: Domain of people/NL communication.
   C: Domain of heat.

Example 3 is a serial mixed metaphor. The "ANGER AS HEAT" metaphor (B as C) acts on the "MIND PARTS AS PERSONS" metaphor (A as B) to directly affect its metaphorical meaning. Therefore, it is not possible to isolate the two metaphors as in example 2. This is an important point: it could be argued that the "ANGER AS HEAT" metaphor applies directly to the contesting by John and not just to the part of John involved in the contesting. Given the definitions above, this would classify example 3 as a parallel mix. However, viewing the utterance as a parallel mix misses a subtle distinction which we wish to capture. The sentence is ambiguous: either one part of John is contesting the verdict and one part is not, and the part contesting is doing so "hotly" or both parts of John are contesting the verdict but only one is doing so "hotly". Our intuitions suggest that the former interpretation is the default one and we will only provide a detailed analysis for this interpretation. However, our treatment is sensitive to such distinctions (as is our computational implementation) and is capable of reasoning about such uncertainties.

Figure 1 represents both examples schematically. Example 2 has two metaphors which act as parallel vehicles on the (literal) tenor domain. However, in example 3, one metaphor (B as C) acts directly on another (A as B). This metaphor is then applied to the tenor. The next section, a computational account dealing with both kinds of mixed meta-

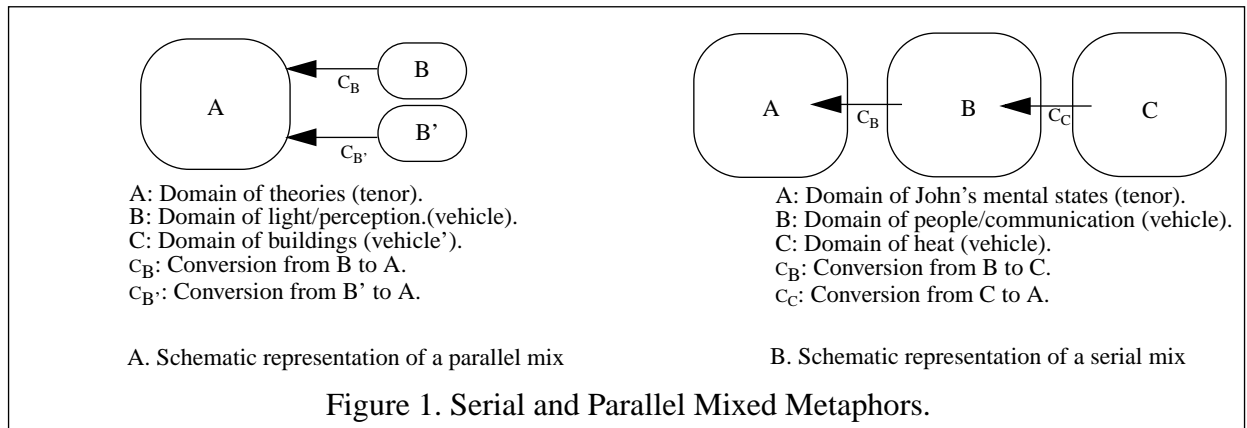

A: Domain of theories (tenor).
B: Domain of light/perception.(vehicle).
C: Domain of buildings (vehicle').
$c_B$: Conversion from B to A.
$c_{B'}$: Conversion from B' to A.

A. Schematic representation of a parallel mix

A: Domain of John's mental states (tenor).
B: Domain of people/communication (vehicle).
C: Domain of heat (vehicle).
$c_B$: Conversion from B to C.
$c_C$: Conversion from C to A.

B. Schematic representation of a serial mix

Figure 1. Serial and Parallel Mixed Metaphors.

phor will be developed based on these distinctions.

## 3. A Computational Account

The work presented here is within the ATT-Meta model of metaphor comprehension. This paper will only detail the concepts relevant to the current work but further details can be found in [Barnden, 1997a].

ATT-Meta is an AI system capable of both simulative reasoning about beliefs and metaphorical reasoning. Reasoning is done by the use of back-chaining rules of inference which allow differing degrees of certainty. Nested reasoning is allowed to facilitate simulation of other agents. In addition, metaphorical reasoning can be nested within simulative reasoning. As the account which follows suggests, such capabilities are central to understanding mixed metaphor. Two types of nested space are maintained: simulation-pretence cocoons and metaphor-pretence cocoons. Simulation-pretence cocoons are used to model the beliefs of other agents. Metaphor-pretence cocoons are a special type of simulation-pretence cocoon where the agent modelled is hypothetical and is assumed to believe the manifested metaphor is literally true.

Knowledge of different domains is encoded in sets of facts and rules which apply to a particular domain. Since metaphors involve a mapping from one domain (the vehicle) to another (the tenor), ATT-Meta uses conversion rules which explicitly map terms from one domain to another. Therefore, any conventional metaphor can be defined by first constructing a set of rules to represent the vehicle domain plus a suitable conversion rule or a small set of such. Understanding proceeds by creating a metaphor-pretence space then mapping any implication to the tenor domain.

An important aspect of the conversion rules is that their degree of certainty can be represented. This for any reasoning about a metaphor to be defeasible and for conflicts between different domains to be handled using general conflict resolution techniques.

So far, the above is common to a number of computational approaches (e.g. [Martin, 1990]). However, ATT-Meta is distinctive in that it licences extensive within-vehicle reasoning in addition to more common, within-tenor reasoning. Rather than simply mapping a correspondence from the vehicle to the tenor and then performing inference to fully understand the connotation of an utterance, ATT-Meta favours extensive inference *prior to mapping*. As we shall see, this gives any conversion rule the important function of filtering out non-relevant parts of a particular metaphor. This is essential for metaphor-pretence spaces to be chained in a sensible manner.

### 3.1. Parallel Mixed Metaphors

As discussed in Section 2, example 2 relies on two familiar metaphors. Considering the former: "IDEAS ARE LIGHT SOURCES", we assume that ATT-Meta is familiar with the metaphor and so knows the following two-way correspondence:

**Conversion Rule:** (See-Believe)
Seeing an Idea corresponds to Believing an idea.

In addition, suppose ATT-Meta believes the following rule concerning the vehicle domain of "seeing":

**(Illumination)** If physical object X is lit by a light source, then X can be seen.

Secondly, we assume that ATT-Meta is familiar with the metaphor "THEORIES AS BUILDINGS" and so, as part of this familiarity, knows the following two-way correspondences:

> **Conversion Rule:** (BuildingsareTheories)
> Buildings correspond to Theories.
>
> **Conversion Rule:** (Instability)
> If X is a theory which is being seen as a building then X is unstable maps to X is implausible.

In addition, that ATT-Meta has the following rule about real buildings:

> **(Stability):** If X is a building and its foundations are weak, then the building is unstable.

Given the above rules, it is possible to infer the connotation that "the critique claimed that the theory was implausible" by the following steps of inference[1]:
1. Construct a metaphor-pretence cocoon where the metaphor "Believing is Seeing" holds.
2. In this space, assert that the critique literally shines light on X where X is the situation described by the remainder of the sentence.
3. Using the Illumination rule, infer that X can therefore be seen.
4. Map that X can be seen to the tenor domain as X can be believed to be true using the Conversion Rule (See-Believe).
5. Construct a metaphor-pretence cocoon where the metaphor "Theories are Buildings" holds.
6. In this space, assert that the theory's foundations were literally unstable.
7. Using the Stability rule, infer that the theory itself was unstable.
8. Map that the theory is implausible using Conversion Rule (Instability).
9. Combine 4 & 8 by substituting the theory's implausibility for X to infer the connotation that the critique licenced the belief that the theory was implausible.

Notice the above analysis allows both instances of metaphor to be reasoned about separately before their literal connotations are combined by a simple substitution. As we shall see in the next section, serial mixes are more complex.

### 3.2. Serial Mixed Metaphors

As discussed in Section 2, example 3 relies on two familiar metaphors. Considering the former: "MIND PARTS AS PERSONS", we assume that ATT-Meta is familiar with the metaphor and so knows the following two-way correspondence:

> **Conversion Rule:** (MindpartsArePersons)
> If person P is viewed as having a part X that is a person, then a motivation/idea of X is a motivation/idea of P.

In addition, some general knowledge is required specifying that when "one person" is mentioned in discourse, then it is reasonable to assume that there is at least more than one person present. This could be analysed as a form of scalar implicature (see, [Hirschberg, 1985; Lee, 1998]). However for the purposes of this paper it is sufficient to use the following defeasible rule in the metaphor domain:

> **SeveralPeople:**
> There is more than one person.

Regarding the second metaphor "ANGER AS HEAT", it is essential to have the following two-way correspondence:

---

1. We make no strong claims as to the psychological ordering of the reasoning steps.

**Conversion Rule:** (HeatisAnger)
Heat proportionally corresponds to emotional anger states.

There is now sufficient information to tackle the serial metaphor. However, unlike the example 2 above, it is not possible to deal with each metaphorical manifestation separately. Instead each metaphorical-pretence cocoon must be chained. Given the above rules, it is possible to infer the connotation that "John had one motivation to angrily challenge the verdict" by the following steps of inference:

1. Construct a metaphor-pretence cocoon M1, where the metaphor "MIND PARTS AS PERSONS" holds.
2. Construct a metaphor-pretence cocoon M2, where the metaphor "ANGER AS HEAT" holds.
3. In M2, assert that PartofJohn1, literally, hotly contested the verdict.
4. Map from M2 to M1, that PartofJohn1, literally contested the verdict angrily, using Conversion Rule (Heat is Anger).
5. Map from M1 to the Tenor domain, that PartofJohn1 corresponds to one motivation of John using the Conversion Rule (MindpartsArePersons).

It is also possible to infer that there is "more than one person in John's head" i.e.:

6. Using the SeveralPeople rule, infer that there is more than one person PartofJohn2
7. Map from M1 to the Tenor domain, that PartofJohn2 corresponds with some other motivation of John.

It can be argued that example 3 implies that the other motivation of John is not to contest the verdict. This is, however, another scalar implicature and therefore, not part of the *metaphorical analysis* of the sentence. However, what ATT-Meta can infer, is that in this context, there is another motivation and this motivation may (or may not) be contrary to the first.

## 4. Further discussion

In Section 3, two types of mixed metaphor were analysed using the same techniques and conceptual structures which have been applied previously to straight metaphors. However, there are some issues particular to mixed metaphor.

It is clear that parallel mixes present less difficulties to any pre-existing theory of metaphor than serial mixes. This is due to the lack of interaction between the two metaphors involved. However, this is not to say that parallel metaphors operate in total isolation. Certain parallel mixes are more common than others. For example, metaphors which refer to abstract entities as physical objects are often mixed with spatial metaphors e.g.:

4. John pushed the ideas to the back of his mind.

Example 4 uses two familiar metaphors: "IDEAS AS PHYSICAL OBJECTS" and "MIND AS ENCLOSED SPACE". However, it is not clear whether such examples are instances of live mixing. There are two reasons for doubt. First, such examples can often be termed *dead mixes* i.e. mixes which have been so conventualised that there is no need for any extra reasoning to combine the two familiar metaphors. This, however, is not to suggest that the individual metaphors are dead, only that the mix is so familiar that any metaphorical reasoning is performed in the same metaphor-pretence cocoon which represents the conventionalised mix of the metaphors.

Secondly, it is not clear whether the level of representation of conceptual metaphors is universal. It is conceivable that two different native speakers may represent the same metaphor with different levels of granularity and in some cases, a manifestation might be mixed to one speaker and straight to another. Therefore, to avoid such issues, we have adopted a position of methodological solipsism [Fodor, 1980] with respect to the particular set of metaphorical concepts assumed and focused on the actual processes and types of data structures involved in reasoning about metaphors. Either case could be made for example 2. It is arguable if there is an actual novel mixing of "IDEAS AS LIGHT SOURCES" and "UNDERSTANDING AS SEEING" or simply a dead mix which could be termed "MENTAL INTERACTION AS VISION".

In our brief references to parallel mixing in earlier work (e.g. [Barnden, 1997a]) we have suggested that such mixes can be handled by having a single metaphorical pretence cocoon, instead of the two assumed in the present paper. In the one-cocoon approach, information in the two vehicle domains can interact (this could be seen as a form of "blending" [Turner & Fauconnier, 1995]). Sometimes such interaction is benign and easy to perform, as in the case of a dead mix, and sometimes fought with conflict (as in example 1). It is a matter of further research to combine the one cocoon and two cocoon approaches to parallel mixing.

In serial mixes, the metaphors strongly interact. If the analysis provided in Section 3.2 is correct, and serial metaphors work by the chaining of one vehicle domain to the other vehicle domain to the tenor, then conversion rules provide an explicit constraint since a sensible mapping is required from the first vehicle to the second vehicle. Therefore, mixing of conventional metaphors is only possible if a conventional mapping exists between the two domains.

By this view, conversion rules act as filters between domains. First, to constrain the types of serial mixed metaphor possible. Secondly, to constrain the types of information transferred since only metaphorical manifestations which make sense in the other metaphor pretence cocoon can be mapped.

In previous work, it has been assumed that generality in conversion rules and mapping is a good thing. However, given this filtering role, specifity is an advantage since it provides strong constraints on mixing. Clearly, within-vehicle reasoning is important here. If more specific conversion rules were favoured then more of the reasoning workload must be performed prior to mapping to the tenor domain.

We have argued previously that contradictions between vehicle and tenor domains can be dealt with using general conflict resolution techniques. Mixed metaphors, however, are more complex. As shown in example 1, a sentence can involve two literally contradictory metaphorical manifestations yet still make sense. This is because only the connotations of each metaphor interact, not the metaphors themselves.

## Conclusions

In this paper, we have described some initial work on mixed metaphors. This paper has argued that both types can be processed using basic AI reasoning techniques which have already been applied to cases of straight metaphor, and in particular, the nesting of simulation and metaphor-pretence cocoons. We have suggested that within-vehicle reasoning plays an important role in mixing so that the connotation of each metaphor can be established prior to mapping to avoid contrary but mixed metaphors from conflicting.